\documentclass[sigconf]{acmart}

\AtBeginDocument{%
  \providecommand\BibTeX{{%
    \normalfont B\kern-0.5em{\scshape i\kern-0.25em b}\kern-0.8em\TeX}}}

\setcopyright{acmcopyright}
\copyrightyear{2022}
\acmYear{2022}

\acmConference[WWW '22]{Proceedings of the ACM Web Conference 2022}{April 25--29, 2022}{Virtual Event, Lyon, France}
\acmBooktitle{Proceedings of the ACM Web Conference 2022 (WWW '22), April 25--29, 2022, Virtual Event, Lyon, France}
\acmPrice{15.00}
\acmDOI{10.1145/3485447.3512144}
\acmISBN{978-1-4503-9096-5/22/04}

\usepackage{todonotes}

\usepackage{caption}
\usepackage{subcaption}
\usepackage{soul}
\usepackage{xspace}

\newcommand{\ang}[1]{\left\langle{#1}\right\rangle}

\newcommand{\es}{\varnothing}
\newcommand{\nat}{\mathbb{N}}

\def\eg{\emph{e.g.}\xspace}

\def\ie{\emph{i.e.}\xspace}

\def\cf{\emph{cf.}\xspace}

\newcommand{\va}[1]{{#1}}
\newcommand{\apy}[1]{{#1}}

\begin{document}

\title{GraphNLI: A Graph-based Natural Language Inference Model for Polarity Prediction in Online Debates}



\author{Vibhor Agarwal}
\affiliation{%
  \institution{Department of Computer Science\\University of Surrey}
  \streetaddress{P.O. Box 1212}
  \city{Guildford}
  \state{Surrey}
  \country{United Kingdom}
  \postcode{GU2 7XH}
}
\email{v.agarwal@surrey.ac.uk}

\author{Sagar Joglekar}
\affiliation{%
  \institution{Department of Informatics\\King's College London}
  \streetaddress{Bush House, Strand Campus\\30 Aldwych, WC2B 4BG}
  \city{London}
  \country{United Kingdom}}
\email{sagar.joglekar@kcl.ac.uk}

\author{Anthony P. Young}
\affiliation{%
  \institution{Department of Informatics\\King's College London}
  \streetaddress{Bush House, Strand Campus\\30 Aldwych, WC2B 4BG}
  \city{London}
  \country{United Kingdom}}
\email{peter.young@kcl.ac.uk}

\author{Nishanth Sastry}
\affiliation{%
  \institution{Department of Computer Science\\University of Surrey}
  \streetaddress{P.O. Box 1212}
  \city{Guildford}
  \state{Surrey}
  \country{United Kingdom}
  \postcode{GU2 7XH}
}
\email{n.sastry@surrey.ac.uk}


\begin{abstract}
Online forums that allow participatory engagement between users have been transformative for public discussion of important issues. However, debates on such forums can sometimes escalate into full blown exchanges of hate or misinformation. An important tool in understanding and tackling such problems is to be able to infer the argumentative relation of whether a reply is supporting or attacking the post it is replying to. This so called  polarity prediction task is difficult because replies may be based on external context beyond a post and the reply whose polarity is being predicted.
We propose GraphNLI, a novel graph-based deep learning architecture that uses graph walk techniques to capture the wider context of a discussion thread in a principled fashion. Specifically, we propose methods to perform root-seeking graph walks that start from a post and captures its  surrounding context to generate additional embeddings for the post. We then use these embeddings to predict the polarity relation between a reply and the post it is replying to. We evaluate the performance of our models on a curated debate dataset from Kialo, an online debating platform. Our model outperforms relevant baselines, including S-BERT, with an overall accuracy of 83\%.

\end{abstract}




\maketitle

\section{Introduction}\label{sec:intro}

The Internet has enabled people to participate in sharing their views, often as comments or posts, about many topics online. Online debates can sometimes become large and acrimonious\apy{, with some escalating into full blown exchanges of hate and misinformation}. As many of these debates concern topics of societal importance, it is \apy{crucial} to be able to model these debates accurately and at scale, so that we can better understand and control phenomena such as the spread of hate~\cite{cinelli2021online,koffer2018discussing}, fake news~\cite{allcott2017social,hanselowski2018retrospective}, how best to moderate political polarisation \cite{bail2018exposure}, and how to break echo chambers by linking appropriate \apy{users} of opposing views~\cite{garimella2017reducing}.

An important task in modelling online debates is to be able to predict whether the reply of one comment to another is \textbf{attacking} (disagreeing) or \textbf{supporting} the post it is replying to. This is called the \textbf{polarity} of the reply \cite{cayrol2005acceptability}. The ability to accurately predict the polarity of  replies in an online debate can allow us to measure properties of the debate, such as how ``controversial'' a discussion is, \eg{} by counting the number of supporting 
\textit{vs.} attacking replies~\cite{boschi2021has}. Perhaps more importantly, if the polarity is known, we can then use techniques from argumentation theory, a branch of artificial intelligence \apy{concerned} with the formal representation and resolution of disagreements \cite{rahwan2009argumentation}, to formally compute which arguments have been attacked and rebutted, and which ones stand unrebutted or are further justified by additional supporting replies.

One obvious approach to predicting the polarity of replies in any debate is to apply natural language processing (NLP) techniques. Such models typically take as input the natural language text of a comment $b$ and the comment $a$ that it is replying to, and output a predicted polarity for whether $b$ is attacking or supporting $a$ (\eg{} \cite{cabrio2013natural,cocarascu2017identifying}). 
However, one shortcoming of such an approach is that it risks loosing crucial information by considering the comments $a$ and $b$ in isolation from the rest of the debate.  

We call the posts $a$, $b$ as the \textit{local} context for the polarity prediction task and the whole discussion thread in which these comments are embedded as the \textit{global} context. This paper asks and answers the question: \textit{Can we improve  performance on the polarity prediction task by incorporating additional context beyond the posts $a$, $b$?}

Typically, discussion threads can be seen as a tree, starting with an original post which forms the root of the tree and each reply such as $b$ creating a directed edge in the tree, from $b$ to its parent node $a$ that it is replying to. We hypothesise that nodes near $a$ and $b$, \eg{} their children, ancestors and siblings in the discussion tree, contain additional information that may help understand whether $b$ is attacking $a$ or supporting it. For example, if other siblings of $b$ (\ie{} children of $a$ other than $b$) are also attacking $a$, then it may be more likely that $b$ is also an attacking reply. \textit{Our key idea is to use graph walk techniques to discover and utilise neighbouring context in a principled fashion.} Our contributions are as follows:

\begin{enumerate}
    \item We compare and contrast several NLP models, including Sentence-BERT~\cite{reimers-2019-sentence-bert}, to establish a baseline for the polarity prediction task. We use data from Kialo, an online debating platform\footnote{See \url{https://www.kialo.com/}, last accessed 17 October 2021.}; this data is in the form of discussion trees where the nodes are arguments submitted to the debate\footnote{We will use the terms ``node'', ``post'', ``comment'' and ``argument'' interchangeably.} and the edges denote which arguments reply to which other arguments, as will be explained in Section~\ref{sec:discussion-trees}.
    \item We propose graph walk techniques that sample discussion trees with the aim of capturing parts of the global context of the online debates, and input the additional nodes sampled into deep learning model along with the local context of the replying argument and the argument being replied to. We find that a weighted root-seeking graph walk works the best in capturing the wider context of the debates on Kialo.
    \item We present and evaluate \textbf{GraphNLI} -- a novel graph-based deep learning architecture to predict the polarity of replies. Our model outperforms baselines including S-BERT, achieving an accuracy of $83\%$. We provide an open source implementation of the model available for public usage\footnote{The model code is available at \url{https://github.com/socsys/GraphNLI} and the dataset is available at \url{https://tinyurl.com/kialo-debates-dataset}.}.
    \item We systematically investigate through ablation studies what features can be helpful in capturing the wider context for the polarity prediction task and show that upstream (or earlier) text, \ie{}, the parent and other ancestor nodes, help the model more than siblings and children replies. Moreover, we find that the importance of ancestor nodes decreases as their distance from the given node increases.
\end{enumerate}


The rest of this paper is structured as follows. Sec.~\ref{sec:background} provides an overview of the polarity prediction task in the context of argumentation theory. In Sec.~\ref{sec:kialo-dataset}, we discuss the details of Kialo dataset. We next consider different ways of incorporating the wider context of discussions in Sec.~\ref{sec:polarity_prediction_model},  and present the GraphNLI architecture. We then evaluate the model in Sec.~\ref{sec:experiments_and_results}, comparing its performance relative to a range of baseline classifiers and conducting an ablation study to better understand which features are important. \apy{In Sec.~\ref{sec:discussion}, we summarize our results, and outline possible future work.}


\section{Background and Related Work}\label{sec:background}

\textbf{Argumentation theory} is a branch of AI that is concerned with the transparent and rational resolution of disagreements (e.g. \cite{rahwan2009argumentation}). Many formal models of argumentation have been devised and studied, starting with \textbf{abstract argumentation theory} \cite{baroni2011introduction,dung1995acceptability,young2018notes}, where the arguments of interest are elements of a set $A$, and a not-necessarily-symmetric binary relation $R\subseteq A\times A$ - the \textbf{attack relation} - represents when two arguments disagree, i.e. $(b,a)\in R$ denotes that argument $b$ disagrees with argument $a$. The directed graph (digraph) $\ang{A,R}$ is called an \textbf{abstract argumentation framework} (AF); this abstracts away from the structure of the arguments and the nature of their disagreements. Resolving these disagreements formally amounts to identifying subsets of arguments $S\subseteq A$ that satisfy various normative properties. \apy{For example,} the property of \textbf{conflict-freeness}, $(S\times S)\cap R=\es$, \apy{formalises the idea of self-consistency because} winning arguments should not attack each other.

Of course, arguments can agree as well as disagree. \textbf{Bipolar argumentation frameworks} (BAFs) enrich AFs with a \textbf{support relation} that models when arguments agree \cite{cayrol2005acceptability}. Formally, a BAF is a digraph $\ang{A,R_{att},R_{sup}}$ where $R_{att}, R_{sup}\subseteq A\times A$ are two disjoint binary relations respectively denoting attack and support between arguments - every edge of a BAF is either attacking or supporting but never both. To determine the winning arguments, there are various ways to transform a BAF into an AF by ``absorbing'' supports into attacks (e.g. \cite{cayrol2013bipolarity}) depending on how supports are interpreted. Following this literature, given an edge in a BAF, we call the status of whether this edge is attacking or supporting its \textbf{polarity}.

Both AFs and BAFs are theoretically interesting and mathematically elegant, and offer a normative and perhaps intuitive way of identifying winning arguments in the presence of agreement and disagreement \cite{cayrol2005acceptability,dung1995acceptability,rahwan2010behavioral}. However, in order to apply these models to reason about online debates, we need to map the concepts identified in these models to their real-life counterparts. \textbf{Argument mining} (e.g. \cite{lawrence2020argument,lippi2016argumentation,cabrio2018five}) is the application of NLP tools to extract arguments and identify their relationships from raw text. Example tasks involve identifying when Tweets from Twitter are well-defined arguments instead of insults, single URLs or pictures (e.g. \cite{bosc2016tweeties}), identifying the claims, their reasons and relationships between claims from clinical trials to inform medical decision making (e.g. \cite{mayer2021enhancing}), or detecting fallacies from the transcripts of the United States Presidential Debates~\cite{villata2021argstrength}. Online debates, such as those on Reddit or Twitter for example, can be converted into the argumentation framework by first defining nodes or arguments corresponding to each well-defined logical claim, and drawing signed edges representing supports or attacks between them. For instance, when a comment $b$ replies to another comment $a$, we may have two nodes $a, b$ linked by an edge representing a support or an attack depending on the polarity relation between them. The result, therefore, is to formally represent an online debate as a BAF.


Many other analyses can be performed once a debate has been represented as a BAF. For instance, based on the polarities of the edges, we may calculate which arguments are justified and which arguments have been rebutted. This could potentially be used to present only the justified arguments as summary to a reader. Previous work has also looked at how the conclusions of a logical reader can change depending on  which parts of a debate they read, thus underscoring the dangers of sampling only parts of a large online debate~\cite{young2018approx,young2020ranking,young2021likes}. Other work has shown how the location of justified arguments can be significantly influenced by whether the debate is acrimonious or supporting~\cite{boschi2021has}. BAFs are thus useful representational tools for modelling online debates, allowing the application of both argument-theoretic and graph-theoretic ideas to gain insights about online discussions.

The polarity prediction task for online debates has been addressed in the argument mining literature.\footnote{See \cite{cabrio2018five} for reviews of the polarity prediction task in other settings, such as persuasive essays or political debates.} An early example is \cite{cabrio2013natural}, which applied textual entailment \cite{bos2006logical,dagan2010recognizing,kouylekov2010open} to predict the polarity of replies on the now-defunct Debatepedia dataset, with a test accuracy of $67\%$. In \cite{cocarascu2017identifying}, long-short-term memory networks were used to classify polarity, achieving $89\%$ accuracy.\footnote{Although this shows better results on the polarity prediction task than what we report here, neither their data nor their framework \apy{were} available for benchmarking.} A more recent overview of the polarity prediction task \cite{cocarascu2020dataset} has provided context-independent baselines of neural network models using a range of learning representations and architectures, and have found an averaged performance of $51\%$ to $55\%$ of these different neural networks across such contexts; these contexts involve online debates on controversial topics such as abortion and gun rights, persuasive essays, and presidential debates. 

In all of the above-mentioned approaches, the inputs to the model are the texts of the replying argument and the argument being replied to, often represented by some appropriate word embedding. Arguably, this is the least amount of information one must input into the model to predict the polarity of the reply. What has not yet been considered is whether it is helpful to input more information. For example, if argument $c$ replies to argument $b$, and $b$ replies to argument $a$, and we would like to predict the polarity relation between $c$ and $b$, then is it useful to design a model that accepts as input the texts of $c$, $b$ \textit{and} $a$? How about if we randomly sample additional ``nearby'' comments? What if we incorporate graph-theoretic features into the input such as in-degree? To the best of our knowledge, these questions have not yet been addressed in the argument mining literature. We thus seek to investigate these questions by measuring whether the incorporation of such features can improve the polarity prediction accuracy.


\section{Kialo Dataset}
\label{sec:kialo-dataset}

\begin{figure*}
\centering
\begin{subfigure}{0.5\textwidth}
  \centering
  \includegraphics[width=\linewidth]{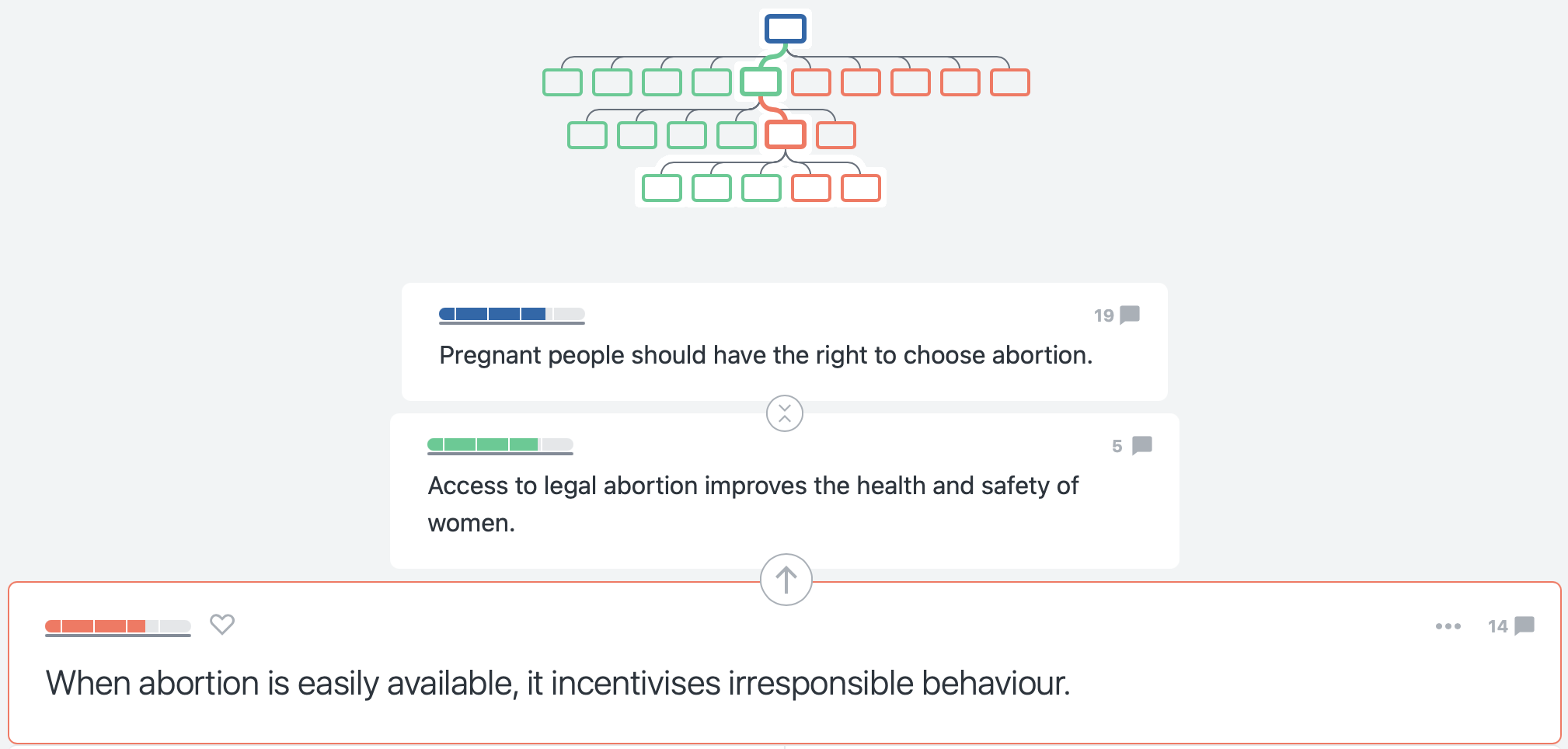}
  \caption{An example of \apy{the arguments made in} a Kialo debate. \apy{The thesis is, ``\textit{Pregnant people should have the right to choose abortion}''. A supporting reply is, ``\textit{Access to legal abortion improves the health and safety of women}''. An attacking reply to this support is, ``\textit{When abortion is easily available, it incentivises irresponsible behaviour}''.}}
  \label{fig:kialo_example}
\end{subfigure}%
\begin{subfigure}{0.4\textwidth}
  \centering
  \includegraphics[width=\linewidth]{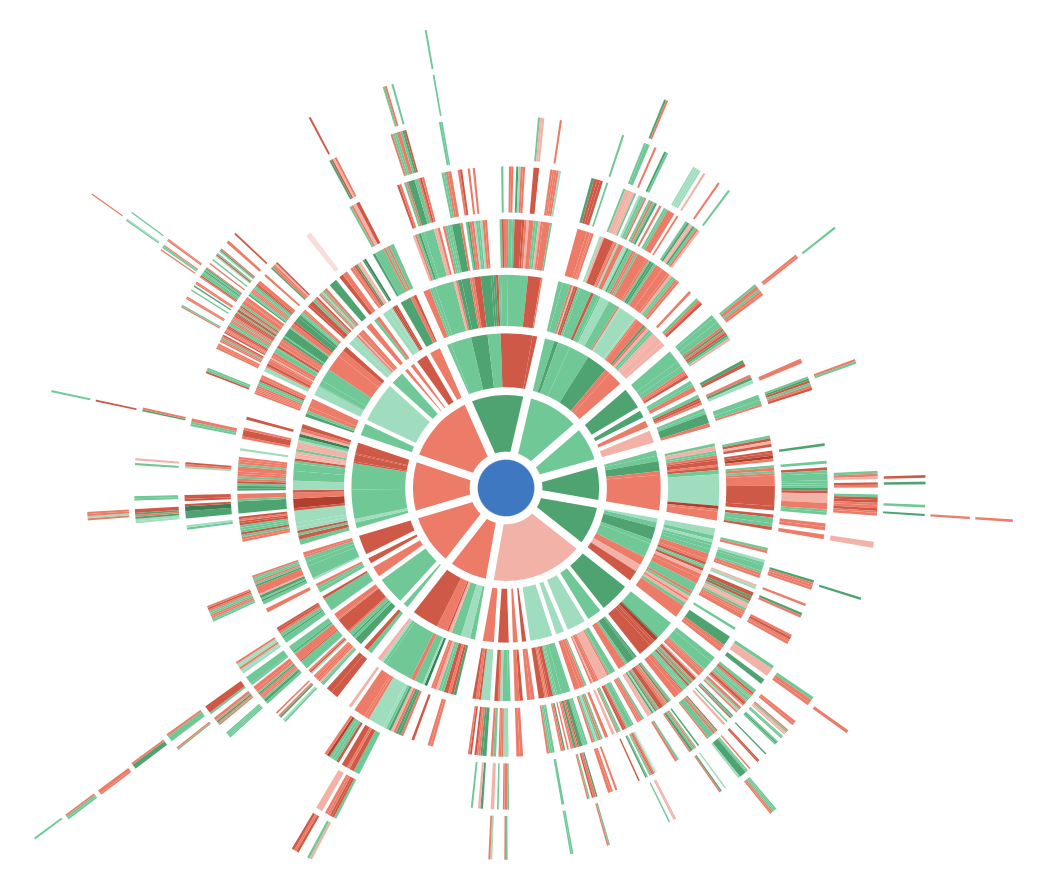}
  \caption{Visualization of this debate's tree}
  \label{fig:kialo_viz}
\end{subfigure}
\caption{Example of a Kialo discussion. Every debate on Kialo would start with a thesis, which in this example is ``\apy{Pregnant people should have the right to choose abortion}''. This thesis can be either supported (shown in green) or attacked (shown in red), and the exchange can go on at multiple levels as seen in Figure \ref{fig:kialo_viz}. \apy{Both figures are taken from \url{https://www.kialo.com/when-abortion-is-easily-available-it-incentivises-irresponsible-behaviour-5637.1340?path=5637.0~5637.1_5637.10160-5637.1340&active=_5637.3912}, last accessed 25 Jan 2022.}}
\label{fig:kialo}
\end{figure*}

Kialo is an online debating platform that helps people ``engage in thoughtful discussion, understand different points of view, and help with collaborative decision-making''.\footnote{Quoted from \url{https://www.kialo.com/about}, last accessed 21 \apy{Oct} 2021.} In this study, we use data from discussions hosted on the Kialo debating platform as used by \cite{boschi2021has}. Figure~\ref{fig:kialo} illustrates an example Kialo discussion. In a Kialo debate, users submit \textbf{claims}. The starting claim of a debate is its \textbf{thesis}. To start a discussion in Kialo, the user creates a thesis along with a tag that indexes the discussion by indicating the content of the discussion. A thesis can have many tags, which increases its visibility to the users. Then users comment on the discussions of their choice, which as the discussion develops, takes the shape of a \apy{directed tree --- this is because each non-thesis claim replies to exactly one other claim.} The dataset contains $1,560$ discussion threads, and is the most complete snapshot of Kialo, as of 28 January 2020.\footnote{\va{We collected data continuously until a change in the Kialo backend in January 2020 which made our crawler obsolete. This was sufficient data to test GraphNLI, so no more data was collected.}} Each discussion thread has data about the tree structure, votes on each argument's impact on the debate it has been submitted to, and the arguments' texts. Further, each reply between arguments is clearly labelled as attacking or supporting. There is also supplemental metadata such as the time of posting, the time of editing, and the author metadata. All discussions crawled from Kialo \apy{thus} have a tree structure with a root node that represents the main thesis and each other node is a reply to its parent which either supports or attacks the parent. On each topic, there is a reasonable amount of debate, with a mean of $204$ and a median of $68$ arguments (standard deviation $463$). Kialo debates are typically balanced, with the vast majority of discussion trees having between $40\%$ to $60\%$ of its edges as supporting edges, with the rest being attacking edges.

Due to Kialo's strict moderation policy, each piece of text submitted to a debate is a self-contained argument that has clear claim backed by reasoning. Thus, each post in Kialo can be taken as a node and directed edges can be drawn based on which post is replying to which other post. The polarity prediction task is to decide whether these edges are attacking or supporting. 

\section{The Polarity Prediction Model}\label{sec:polarity_prediction_model}

As stated in Section \ref{sec:background}, polarity prediction is an important task in argument mining. It aims to identify the argumentative relations of attack and support between natural language arguments by classifying pairs of text accordingly, where in our case such texts are comments submitted to online debates, and one text is replying to another text. Various deep learning models have been used in the literature to perform the polarity prediction task. However, they usually consider a pair of texts for the polarity prediction. In this section, we propose a novel graph-based deep learning architecture that not only considers as input the pair of texts, but also \apy{systematically} captures the context of nearby comments via graph walks.

\subsection{Representing Debates as Discussion Trees}\label{sec:discussion-trees}
 For every online debate \textit{D} on Kialo, we construct a tree structure, where a node represents a post and a \textit{directed} edge from a node to its parent, the post it is replying to. Each such edge has an associated label,  \apy{\textit{support} or \textit{attack}},  depending upon whether the comment is \apy{respectively for or against} its parent comment. The root node of this discussion tree represents the thesis (topic) of a debate. 

\subsection{GraphNLI Architecture}

\begin{figure}[h]
  \centering
  \includegraphics[width=\linewidth]{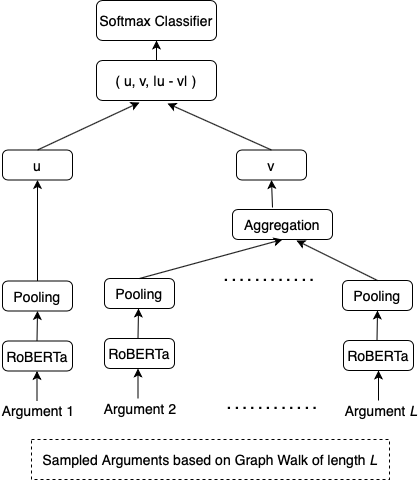}
  \caption{GraphNLI Architecture}
  \label{fig:graphnli-arch}
\end{figure}

We propose a novel graph-based deep learning architecture which captures both the local and the global context of the online debates through graph-based walks. The GraphNLI architecture is shown in Figure \ref{fig:graphnli-arch}, and will be explained in Section \ref{sec:model_overview}.

\subsubsection{\textbf{Capturing Global Context through Graph Walks}}
\label{sec:graph-walks}

Our GraphNLI model captures the global context of online debates through graph-based walks. A walk is defined as a sequence of nodes traversed from a given node in a tree. \textit{We choose to ignore the direction of the edges in these walks}, to enable capturing context such as siblings of the reply node (reachable by going up to the parent node and then back down again to the sibling). Walk length \apy{$L\in\nat$} is the maximum allowed length after which the graph walk terminates. We propose the following strategies for graph walks in discussion trees to capture the global and neighbouring context of a given argument (node).

\textbf{Weighted Root-seeking Graph Walk}: In a discussion tree, a root-seeking graph walk is a walk starting from a given node \apy{\textit{up}} towards the root of the discussion tree. In a tree, there is exactly one path that goes upwards to the root of the tree from a given node. Depending on the walk length $L$, the graph walk terminates irrespective of whether or not it reaches the root node. We experiment with different walk lengths and find that $L = 5$ gives the best results \apy{(see Section \ref{sec:experiments_and_results})}. The graph walk is \apy{\textit{weighted}}, which means that the contributions of the ancestor nodes in the node's embedding vector \va{are discounted by a factor of $\gamma^k$, where $k$ is the distance of the node from the given node and $\gamma$ is the \textit{discount parameter}}. Therefore, the highest weight is given to the node's parent, then a discounted weight to the parent's parent, and so on. This means the closer the node is to the starting node, the higher its weight will be, and the more it will contribute as an input towards predicting the polarity.

\begin{example}\label{eg:weighted_root_seeking_walk}
Suppose we have a section of a discussion tree with arguments $a_0$, $a_1$, $a_2$, $a_3$, $a_4$ and $a_5$, where the edges are $(a_k, a_{k+1})$ for $0\leq k\leq 4$, and $a_{k+1}$ is the parent of $a_k$. As $L=5$, a weighted root-seeking graph walk \apy{from $a_0$} will sample arguments $a_1$ to $a_5$. As for their weights, suppose we use a discount parameter $\gamma=0.5$, we compute the weight of $a_k$ as $\gamma^k$. Therefore, the weight of $a_0$ is $1$, the weight of $a_1$ is $0.5$, the weight of $a_2$ is $0.25$, and so on.
\end{example}

\begin{example}\label{eg:weighted_root_seeking_walk2}
Suppose we have a section of a discussion tree with arguments $a_0$, $a_1$, $a_2$, where the edges are $(a_k, a_{k+1})$ for $0\leq k\leq 2$, and $a_{k+1}$ is the parent of $a_k$. Further, assume $a_2$ is the root of the tree. As $L=5$, a weighted root-seeking graph walk from $a_0$ will sample arguments $a_1$ and $a_2$, and no more as $a_2$ is the root of the tree itself.
\end{example}

\textbf{Biased Root-seeking Random Walk}: This is a random walk starting from a given node in the tree and is biased towards the root. 
From a given node, each immediate neighbor is assigned a probability of being selected by the random walk. Since the discussion thread is a tree, each node has one parent and zero or more children. To bias the walk towards the root (or thesis node), we assign a probability $p>0.5$ to the random walk selecting the parent of the given node as the next step in the walk. Recall that we ignore the direction of edges in these walks, allowing the walk to traverse downwards as well as upwards in the discussion tree. Therefore, the remaining probability, $1-p< 0.5$, is divided equally among all the children of the given node. The walk length \apy{$L\in\nat$} determines the maximum length of this random walk, that is, the number of nodes to be visited from the given node until the random walk terminates. After experimentation, we find $L = 4$ to be optimal \apy{(see Section \ref{sec:experiments_and_results})}. 

The random walk is a different way of sampling the neighbouring nodes as a means of incorporating the surrounding \apy{global} context for predicting the polarity. \va{Note that there is no guarantee that the parent node will be visited in the random walk. By choosing the probability $p$, we can directly affect the probability of visiting the parent. Empirically, we find that $p=0.75$ gives the best results. Even if the parent is not visited, there is likely to be information in the surrounding nodes that still helps predict the polarity of the parent-child relationship. For example, if the majority of children nodes replying to the parent are attacking (perhaps because the parent post makes a controversial or wrong statement), knowing the sibling context may help predict the polarity of a reply}.

\begin{example}
Suppose we have a section of a discussion tree with nodes $a_0$, $a_1$, $a_2$, $a_3$ and $a_4$, with edges $(a_1,a_0)$, $(a_2,a_1)$, $(a_3,a_1)$ and $(a_4,a_1)$, such that for $(a_i,a_j)$, $a_j$ is the parent of $a_i$. Let $p=0.75$, then $1-p=0.25$. A random walk starting at $a_1$ will have probability $0.75$ moving to $a_0$ next. Similarly, starting from $a_1$, there is a  probability \apy{$\frac{0.25}{3}=\frac{1}{12}$} of moving to any of $a_2$, $a_3$, or $a_4$ next.
\end{example}

Also, the same node can be visited multiple times in a random walk, especially when the graph walk chooses one of the children nodes:
\begin{example}
Suppose we have a section of a discussion tree with nodes $a_0$, $a_1$, $a_2$, $a_3$ and $a_4$, with edges $(a_1,a_0)$, $(a_2,a_1)$, $(a_3,a_1)$ and $(a_4,a_2)$, such that for $(a_i,a_j)$, $a_j$ is the parent of $a_i$. Let $p=0.75$, then $1-p=0.25$. Assume that a random walk, starting at $a_1$, moves to $a_2$ with probability \apy{$\frac{0.25}{2}=\frac{1}{8}$}. Now from $a_2$, \apy{the} random walk can either move to $a_1$ \apy{\textit{again}} with probability $0.75$ or \apy{to} $a_4$ with probability $0.25$.
\end{example}

At the end of a graph walk for each node, we obtain at most \textit{L} arguments which are either parents or neighbouring responses of a given node. We input these sets of arguments into our GraphNLI model as described in the next section.

\subsubsection{\textbf{Model Overview}}\label{sec:model_overview}

Our model is inspired by S-BERT~\cite{reimers-2019-sentence-bert}; its architecture is shown in Figure~\ref{fig:graphnli-arch}. Firstly, each of the \textit{L} arguments sampled by the graph walk or the random walk is input into the RoBERTa model~\cite{liu2019roberta} to get their corresponding embeddings and then, a mean-pooling operation, that is, \apy{calculating the} mean of all the output vectors, is applied to derive a fixed-sized sentence embedding for each argument. \va{The starting node in a graph walk is a \textit{point-of-interest} (PoI) node. Let $u$ denote the sentence embedding corresponding to the PoI node. Let $v$ denote the aggregated embedding from its neighbors (including parent) sampled by the graph walk starting from the PoI node. These $u$ and $v$ embeddings together are used in GraphNLI model to predict the polarity.} We experiment with three aggregation strategies: \apy{\textit{summation}, \textit{average} and \textit{weighted average}}. As stated in Section~\ref{sec:graph-walks}, nodes sampled by a weighted root-seeking walk are weighted in the descending order from the PoI node up to the root. The resultant sentence embeddings $u$ and $v$ are concatenated with element-wise difference $|u - v|$ to get the final embedding vector, which is then fed into a softmax classifier for the polarity prediction task.

In order to fine-tune BERT, we make the GraphNLI model end-to-end trainable to update weights during backpropagation such that the produced sentence embeddings are semantically meaningful for the downstream prediction tasks.

\section{Experiments and Results}\label{sec:experiments_and_results}

In this section, we describe our experimental setup for training the GraphNLI model, the results obtained, and a detailed ablation study to better understand different aspects of the model.

\subsection{Dataset Preprocessing}

As described in Section~\ref{sec:kialo-dataset}, we use data from online debates conducted on Kialo in order to train and evaluate our model. All discussions crawled from Kialo have a tree structure with a root node that represents the main thesis and each other node is a reply to its parent which either supports or attacks the parent.

As discussed in Section~\ref{sec:discussion-trees}, we represent Kialo debates as undirected discussion trees. Each edge can have a positive or negative sign, indicating support or attack respectively. \apy{As mentioned in Section \ref{sec:kialo-dataset}}, Kialo debates are typically balanced, with the majority of discussion trees having a fraction of supportive replies between 0.4 and 0.6 with the rest being attacks; \apy{this justifies our use of accuracy as the metric to evaluate our polarity classifiers (see Section \ref{sec:evaluation}). We randomly sample $80\%$ of the Kialo debates into a training set with the remainder serving as a test set}. Overall, the \apy{training} set contains of $259,499$ arguments (replies) in total, while the test set contains $64,874$ arguments in Kialo debates.

\subsection{Training Details}

After preprocessing the Kialo dataset, we use \apy{the} graph walk techniques described in Section~\ref{sec:graph-walks} to capture the \apy{neighbourhood and parent contexts} for each of the nodes and feed them into our GraphNLI model.

In \apy{the} case of the weighted root-seeking graph walk, we assign weights to the nodes in the graph walk progressively. We define a discount parameter $\gamma = 0.75$ such that the nodes are weighted in the form of $\gamma^k$, where $k$ is the distance of the node from the given node. So, the \apy{largest} weight $\gamma^1 = 0.75$ is assigned to the node's parent, and then $\gamma^2$ is assigned to the parent's parent and so on in the graph walk with $L$ nodes.  Therefore, the influence of the ancestor nodes decreases \apy{exponentially} as we move farther away from the given node towards the root in a graph walk.

We fine-tune GraphNLI model with a softmax classifier objective function and cross-entropy loss for four epochs. We use a batch-size of 16, Adam optimizer with learning rate $2\times 10^{-5}$, and a linear learning rate warm-up over $10\%$ of the training data.

\subsection{Baselines}
\label{sec:baselines}

We compare GraphNLI with the following relevant baselines on the classification accuracy.

\textbf{Bag-of-Words with Logistic Regression}: A bag-of-words (BoW) model that uses unigram features as input obtained from the arguments/replies in online debates. Then the parent and the child BoW embeddings are concatenated and fed into a Logistic Regression classifier with L2 regularization trained for 100 epochs.

\textbf{Prompt embeddings (Rhetorics) with Logistic Regression}: The prompt embedding model~\cite{zhang2017asking} infers a vector representation of utterances / arguments in terms of the responses that similar utterances tend to prompt, as well as the rhetorical intentions encapsulated by such utterances / arguments. Once the embedding vectors for the arguments are obtained, the parent and child embeddings are concatenated and input to a Logistic Regression classifier for the final polarity prediction task. The classifier is trained for 100 epochs with L2 regularizaton and cross-entropy loss.

\textbf{Sentence-BERT}: S-BERT~\cite{reimers-2019-sentence-bert} is a modification of a pre-trained BERT transformer network~\cite{devlin2018bert} to derive semantically meaningful sentence embeddings. We use the S-BERT architecture with classification objective function and input the sentence pairs (parent and child arguments) into the model to get their sentence embeddings. Later on, these embeddings are concatenated and fed into a softmax classifier. The S-BERT model is fine-tuned on Kialo training dataset for 4 epochs with a batch-size of 16, the Adam optimizer, and cross-entropy loss function.

\textbf{Non-trainable BERT embeddings with graph \apy{walks} and Multi-layer Perceptron}: For each of the arguments in online debates, their embeddings are derived using a pre-trained BERT model and using CLS-token embeddings. Using different graph walk techniques as described in Section~\ref{sec:graph-walks}, various neighborhood siblings and parent nodes are sampled for each node, and using their node embeddings, a resulting aggregated embedding is formed using an average aggregation function. These node embeddings are then fed into a multi-layer perceptron (MLP) with two layers and with a softmax objective function for polarity prediction. The initial BERT embeddings are non-trainable. We train the MLP for 50 epochs or until the model converges on Kialo training dataset with batch-size of 16, and the Adam optimizer.

\subsection{Model Evaluation}\label{sec:evaluation}

\begin{table}
  \caption{Accuracy scores of different models trained on Kialo dataset for polarity prediction, discussed in Section \ref{sec:evaluation}.}
  \label{tab:results}
  \resizebox{\columnwidth}{!}{
  \begin{tabular}{lc}
    \toprule
    \textbf{Model} & \textbf{Accuracy (\%)} \\
    \midrule
    Bag-of-Words + Logistic Regression & 67.00 \\
    Prompt Embeddings + Logistic Regression & 61.20 \\
    Sentence-BERT with classifier layer & 79.86 \\
    BERT Embeddings: Root-seeking Graph Walk + MLP & 70.27 \\
    \midrule
    GraphNLI: Root-seeking Graph Walk + Sum & 80.70 \\
    GraphNLI: Root-seeking Graph Walk + Avg. & 81.96 \\
    GraphNLI: Root-seeking Graph Walk + Weighted Avg. & \textbf{82.87} \\
    GraphNLI: Biased Root-seeking Random Walk + Sum & 79.95 \\
    GraphNLI: Biased Root-seeking Random Walk + Avg. & 80.44 \\
    \bottomrule
  \end{tabular}
  }
\end{table}

We evaluate the performance of GraphNLI model on the test set of Kialo dataset. Since the polarity prediction task is a binary classification problem and that the datasets are roughly balanced between both classes (attacks and supports), we use accuracy as the evaluation metric. We train models with five different random seeds and report their average performances. Table~\ref{tab:results} shows the accuracy scores of different models trained on Kialo train set for the polarity prediction task. The baseline \apy{model}, Bag-of-Words embeddings with Logistic Regression\apy{,} achieves an accuracy of $67\%$, while Prompt embeddings (Rhetorics) with Logistic Regression achieves just $61.20\%$ accuracy. The sentence-BERT model trained on Kialo dataset achieves an accuracy of $79.86\%$. The initial MLP model with non-trainable BERT embeddings and root-seeking graph walk achieves an accuracy of $70.27\%$ which is even worse than the Sentence-BERT.

Our model, GraphNLI with root-seeking graph walk and averaging node embeddings in the graph walk to get the aggregated node embeddings achieves an overall accuracy of $81.96\%$, whereas, the model achieves even better accuracy of $82.87\%$ using weighted average node embeddings. Using biased root-seeking random walk and average node embeddings, GraphNLI achieves an accuracy of $80.44\%$. Clearly, all the variants of GraphNLI model achieve better accuracy scores than all the baselines including sentence-BERT. GraphNLI with root-seeking graph walk and weighted average node embeddings achieves the highest accuracy overall. This shows that global context of the online debates or discussions along with the local context of the argument pairs indeed helps in improving the performance of the model.

The best performing variant, GraphNLI with root-seeking graph walk and weighted average node embeddings, shows that having context of the upstream arguments in the online debates helps the model in predicting argumentative relations of support and attack. Also, weighted average aggregation gives higher weights to the arguments near to the given argument pair in the discussion tree whose polarity needs to be predicted, and progressively reduces the weights when the graph walk moves towards the root.


\subsection{Ablation Study}\label{sec:ablation_study}

We have demonstrated superior performance of the GraphNLI architecture \apy{with respect to various baselines in Table \ref{tab:results}}. In this section, we perform an ablation study and discuss different aspects of the GraphNLI model and intuition behind the choices in order to gain a better understanding of the model.

First, we evaluate different kinds of graph walks by feeding the resultant embeddings obtained with the \textit{average} aggregation strategy into our model and compare their accuracy scores. As shown in Table~\ref{tab:results}, all the graph walks with GraphNLI architecture perform better than the S-BERT model significantly. S-BERT just considers the argument pairs (node and its parent embeddings), whereas through graph walks, GraphNLI considers the global context of discussion trees by exploring parents and neighborhoods of a node. Therefore, the global context with the local context of the ongoing discussions indeed helps in predicting the polarities. On comparing different graph walks themselves, we find \apy{the} Root-seeking Graph Walk\apy{s performing} better than the Biased Root-seeking Random Walk\apy{s} with $1.52\%$ higher accuracy. It shows that upstream text in the discussions (parent and other ancestor nodes) indeed helps the model more than the sibling and child nodes \apy{possibly} obtained with the biased root-seeking random walk. Intuitively, a person's reply can be influenced by the on-going discussion in the upstream text, but \apy{cannot} be influenced by future replies (child nodes) or sibling \apy{(parallel) replies}.

We evaluate different aggregation strategies (\textit{summation}, \textit{average} and \textit{weighted average}) to aggregate the node embeddings of the neighbouring nodes using a \apy{Root-seeking Graph Walk}. As shown in \apy{Table~\ref{tab:results}}, \apy{the} weighted average aggregation function performs better than \apy{the} summation \apy{and} average strategies. This shows that influence of the upstream text (ancestor nodes) decreases as the graph walk moves away from the given node towards the root. Hence, ancestor nodes \apy{cannot} be weighted equally but instead, progressively in the decreasing order of their distance from the given node.

\begin{table}[h]
  \caption{Accuracy scores of GraphNLI model trained on Kialo dataset with different concatenation techniques using weighted average aggregation, discussed in Section \ref{sec:ablation_study}.}
  \label{tab:ablation-results}
  \begin{tabular}{lc}
    \toprule
    \textbf{Concatenation} & \textbf{Accuracy (\%)} \\
    \midrule
    $(u, v)$ & 76.78 \\
    $(u, v, u*v)$ & 82.05 \\
    $(u, v, |u-v|)$ & \textbf{82.87} \\
    $(u, v, |u-v|, u*v)$ & 82.38 \\
    \bottomrule
\end{tabular}
\end{table}

We also evaluate different  methods for concatenating a node's embedding \textit{u} with the aggregated embedding of its neighbours \textit{v} obtained using the root-seeking graph walk. The impact of \apy{the} concatenation method on the model's performance is significant. As depicted in Table~\ref{tab:ablation-results}, \apy{the} concatenation of $(u, v, |u-v|)$ works the best. As reported by~\cite{reimers-2019-sentence-bert}, adding the element-wise \apy{multiplication} $u * v$ decreased  performance. Element-wise \apy{absolute} difference $|u - v|$ which measures the distance between the two node embeddings is an important component.

Our initial non-end-to-end trainable model as discussed in Section~\ref{sec:baselines} in which we keep the node embeddings obtained from the BERT model fixed, performs even worse than the Sentence-BERT. This throws light on the importance of end-to-end training of the model for fine-tuning on specific tasks. After end-to-end training, the model outputs node embeddings that are rich in context suitable for downstream tasks like polarity prediction.

\section{Conclusions and Future work}
\label{sec:discussion}


In this paper, we demonstrated a novel model, GraphNLI, that quantifies the polarity of an online interaction by building a representation that captures its content and context. \va{GraphNLI  derives inspiration from S-BERT, but it is novel in the sense that it captures the context by sampling a discussion tree of a discourse using different strategies based on graph walks (Section \ref{sec:graph-walks}). Empirically, we found that a Root Seeking Graph Walk with a weighted average aggregation of the ancestors' contexts is the best strategy in terms of classification accuracy (Table \ref{tab:results}).} This strategy addresses the shortcomings of previous approaches that only capture the local context features, such as the reply and the post it is replying to. We also showed through an ablation study that information from parent and other ancestor nodes provide more relevant contextual information than siblings and children of the reply node whose polarity is being determined. Furthermore, the importance of ancestor nodes decreases as the distance from the reply node increases.  


A framework for polarity prediction, such as GraphNLI, can have a number of applications for Computational Social Science (CSS):

\noindent\textbf{Understanding conversation health}: Online discussion forums provide a great opportunity for creative and socially positive interactions, such as peer-support for long-term medical problems~\cite{joglekar2018online,panzarasa2020social}. On the other hand, many forums have unfortunately become a medium for rampant misinformation~\cite{kumar2016disinformation} and hate~\cite{gagliardone2015countering}. As such, identifying and promoting ``healthy'' conversations has been identified as an important priority by many (\eg{} Twitter~\cite{twitter2022healthyconv}). Once the polarity of posts is known, it can be used to develop  conversation health metrics such as whether a conversation is supporting or acrimonious (\eg{}, \cf{} ~\cite{boschi2021has}). 

\noindent\textbf{Hate speech}: Hate speech on online forums is a  common~\cite{cinelli2021online} challenge, including in nationally important conversations between citizens and their elected representatives~\cite{pushkal21MPHate}. In other cases, some members of a discussion can be unfairly targeted, for example, misogyny is understood to be an important problem on online forums such as Reddit~\cite{guest2021expert}. Inferring argument polarities at scale can help platforms to detect such problems before they spiral out of control (for example, highly attacking comments towards female participants can be a possible indicator of potential misogyny). 

\noindent\textbf{Detecting filter bubbles}: Democratic conversations on news and social media sites can exhibit partisan tendencies~\cite{NRSWWW2018b,agarwal2021under,karamshuk16slant,NRSWWW2020}. This can lead to filter bubbles:  two (or more) parallel conversations about the same topic, with each conversation consisting of posts largely agreeing with other posts in that conversation, and yet having a large amount of disagreement with the other conversations happening in parallel.
 Frameworks like GraphNLI could help detect filter bubbles by quantifying agreeability in conversations: \eg{} if we find that posts reachable from each other also agree with each other (\textit{i.e.,} are supporting), and yet if an imaginary edge is induced between posts in different parts of a conversation (or a different discussion thread), we find that the imaginary edge would be an attack edge, this could be indicative of a filter bubble.
 
\noindent\textbf{Understanding multiple viewpoints in online debates}: A novel way to mitigate problems such as filter bubbles in large online debates is to present readers with a balanced sample of all the justified arguments representing the multiple important viewpoints. Once an argumentation framework has been induced from a discussion and the attack/support relation between posts has been established, it is possible to use the tools of argumentation theory to compute the ``winning'' arguments from all sides of a debate~\cite{young2018approx,young2020ranking,young2021likes}:  arguments which have not been refuted and are left standing as valid viewpoints (whether one may agree with them or not).

In future work, we hope to apply GraphNLI to some of the above.

Our work partly fixes the gap of inferring polarity from the content and context of an online discourse. However, several questions remain to be answered to make it relevant for wider usage. For instance, we have demonstrated and evaluated our approach on Kialo, a tightly moderated online debating platform. We would, however, like to test its validity on much noisier, weakly moderated discourses, such as those found on Reddit.\footnote{\apy{\url{https://www.reddit.com/}, last accessed 22 Jan 2022.}}. Other forums, such as BBC's \textit{Have Your Say}?\footnote{\apy{See the comments of, e.g. \url{https://www.bbc.co.uk/news/uk-43253389\#comments}, last accessed 22 Jan 2022.}} do not have an explicit threaded reply structure, requiring us to \textit{infer} from the text of a reply which other post it is replying to, prior to applying GraphNLI's graph walk techniques. In this less restrictive user interface, posts may refer to multiple other posts, which in turns means that the reply graph is no longer a tree. However, we note that discussions that are not trees pose no limitations to our graph walk techniques (Section \ref{sec:graph-walks}) because there would only be more contexts for the graph walk to sample. 

\bibliographystyle{ACM-Reference-Format}
\bibliography{WWW2022}

\end{document}